\documentclass{article}
\usepackage[main, nonanonymous, final]{sty_file}

\usepackage[utf8]{inputenc} 
\usepackage[T1]{fontenc}    
\usepackage{hyperref}       
\usepackage{url}            
\usepackage{booktabs}       
\usepackage{amsfonts}       
\usepackage{nicefrac}       
\usepackage{microtype}      
\usepackage{xcolor}         

\usepackage{microtype}
\usepackage{graphicx}
\usepackage{subcaption}
\usepackage{booktabs}
\usepackage{hyperref}
\usepackage{url}
\usepackage{amsmath,amssymb,amsfonts,amsthm}
\usepackage{bbm}
\usepackage{textcomp}
\usepackage{wrapfig}
\usepackage{xcolor}    
\usepackage[table]{xcolor}
\usepackage{amsfonts}     
\usepackage{nicefrac}    
\usepackage{enumitem}
\usepackage{multirow,tabularx}
\usepackage{inconsolata}
\usepackage{epsfig}
\usepackage{makecell}
\theoremstyle{definition}
\usepackage{tcolorbox}

\usepackage[capitalize,noabbrev]{cleveref}

\title{MoG: Mixture of Experts for Graph-based Retrieval-Augmented Generation}

\author{
  Zheng Yuan\textsuperscript{\dag}, Chuang Zhou\textsuperscript{\dag}, Linhao Luo\textsuperscript{\ddag}, Siyu An\textsuperscript{\S}, Di Yin\textsuperscript{\S}, Xing Sun\textsuperscript{\S}, \textbf{Xiao Huang}\textsuperscript{\dag}\\
  \textsuperscript{\dag}The Hong Kong Polytechnic University, \textsuperscript{\ddag}Monash University, \textsuperscript{\S}Tencent Youtu Lab\\
   \small{\texttt{\{yzheng.yuan,chuang-qqzj.zhou\}@connect.polyu.hk,xiaohuang@comp.polyu.edu.hk}}
  }
\begin{document}

\maketitle

\begin{abstract}
Retrieval-augmented generation is intensively studied to ground large language models on external evidence. However, retrieving from a unified knowledge base could inevitably introduce irrelevant information that may mislead generation for complex reasoning. Inspired by the conditional computation of mixture of experts (MoE), where a router sparsely selects specialized experts alongside shared ones for each input, we propose \textbf{M}ixture \textbf{o}f experts for \textbf{G}raph-based Retrieval-Augmented Generation, i.e., \textbf{MoG}. It organizes knowledge into two core components: (i) diverse, always-accessible hub graphs that encode semantically and structurally central knowledge and provide contextual clues for expert activation, and (ii) sparsely activated expert graphs that contain domain-specific evidence. MoG first accesses hub graphs to identify general evidence and derive contextual clues. Then, a topology-aware router dynamically activates a limited set of expert graphs conditioned on the query, thereby confining retrieval to a focused evidence subspace. Extensive experiments on challenging benchmarks show that MoG consistently outperforms strong baselines, with over 20\% relative improvement on MuSiQue. Our code is available in \url{https://github.com/DEEP-PolyU/MoG}.
\end{abstract}

\section{Introduction}
Large language models (LLMs) have demonstrated remarkable capabilities in generation and reasoning. However, they fundamentally rely on parametric knowledge, which often leads to hallucinations and factual errors. This limitation is particularly acute in complex reasoning tasks, such as multi-hop question answering, that require integrating multiple pieces of evidence.
Retrieval-Augmented Generation (RAG) has been intensively studied to ground LLMs with external corpus~\cite{feng2024don,ke2025survey}, providing explicit evidence to improve factual accuracy~\cite{dong2024modality,jiang2023active,lewis2020retrieval}. Despite its success, RAG faces persistent challenges when applied to complex scenarios that require multi-hop reasoning~\cite{dong2023hierarchy,luo2025gfm}.

\begin{figure}[!t]
	\centering
        \includegraphics[width=1.0\linewidth]{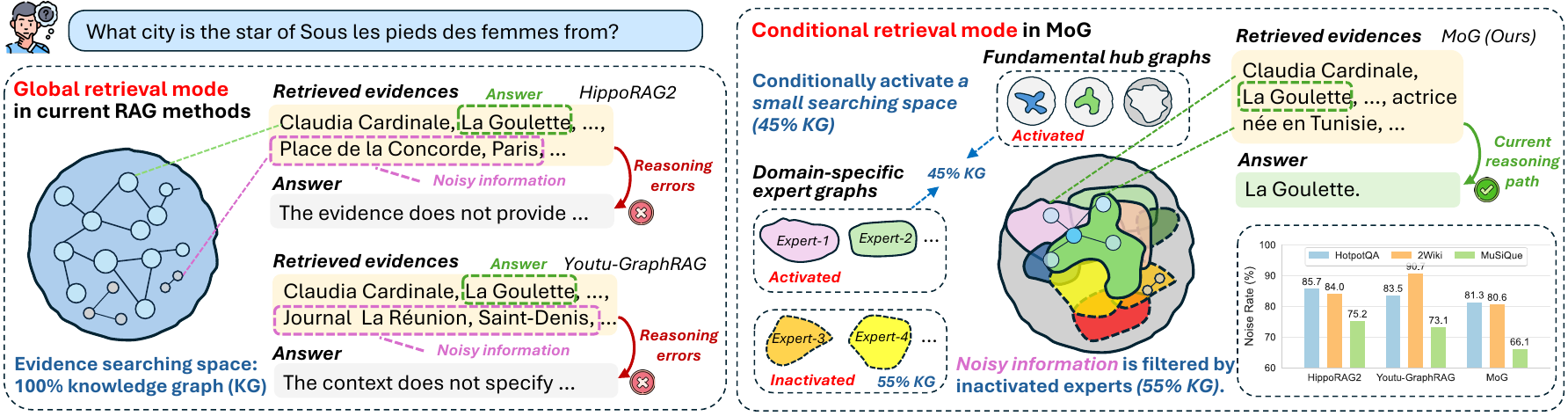}
        \caption{In global retrieval mode, noisy information retrieved alongside correct answers can interfere with the LLM's reasoning. The proposed MoG employs conditional retrieval to narrow the search space, thereby reducing the retrieval noise and providing precise evidence.} 
        \label{fig: intro figure}
\end{figure}

To better capture dependencies among different concepts, recent work has explored structured retrieval beyond flat document ranking in the form of graphs, i.e., GraphRAG. These methods organize external knowledge using hierarchical chunking \cite{sarthi2024raptor}, concept graphs \cite{guo2024lightrag,gutierrez2024hipporag,zhao20252graphrag}, or hybrid structures combining symbolic and textual representations \cite{dong2025youtu,gutierrez2025rag}. While these approaches improve structural awareness and connectivity modeling, they rely on global search over a unified knowledge base for each query, as illustrated in Figure~\ref{fig: intro figure}. Consequently, the search space remains extremely large while unrelated neighbors and noisy reasoning paths could be inevitably introduced during retrieval. This lack of adaptivity to the query often yields evidence sets that are semantically related yet contain substantial irrelevant content, thereby degrading the precision of downstream answer generation.

Inspired by Mixture-of-Experts (MoE) models~\cite{mu2025comprehensive,zhou2022mixture}, which employ conditional computation to activate only a small subset of specialized experts for each input, we adopt an analogous paradigm for GraphRAG. By enforcing sparse, targeted specialization (e.g., expert selection), this approach substantially reduces the effective search space while preserving access to general contextual knowledge. A router governs sparse expert activation, and shared components provide broad coverage, thereby balancing expressiveness and efficiency by conditioning retrieval and computation on the input.

However, leveraging MoE into GraphRAG is promising but challenging. \textbf{Challenge 1: Knowledge organization.} Real-world knowledge is inherently broad and interconnected, and naively splitting a unified knowledge base into several experts can break the multi-hop connectivity required to reach the evidence for complex reasoning. \textbf{Challenge 2: Reliable routing.} While sparse activation can substantially reduce the effective search space, an inappropriate router strategy may instead confine retrieval to an irrelevant region, thereby preventing access to the evidence needed for the answer. These challenges highlight the need for a framework that organizes broad knowledge without sacrificing multi-hop connectivity, while using reliable clues to guide sparse expert activation.

To address these challenges and bring Mixture-of-Experts style conditional retrieval into GraphRAG, we propose \textbf{M}ixture \textbf{o}f experts for \textbf{G}raph-based Retrieval-Augmented Generation (\textbf{MoG}). 
For Challenge 1, MoG first constructs a knowledge graph from an external corpus and organizes it into two complementary components: 
\textbf{hub graphs}, which are always accessible and contain general, semantically and structurally central knowledge that frequently bridges multiple domains via multi-hop connections; 
and sparsely activated \textbf{expert graphs}, which encode domain-specific evidence; entities may appear in multiple hub and expert graphs to reflect the overlapping nature of real-world knowledge.
For Challenge 2, MoG introduces \textbf{a topology-aware router}, a non-parametric component that operates purely on graph topology and entity memberships. 
Using contextual clues retrieved from hub graphs based on the query, the router selects a small subset of expert graphs, thereby confining retrieval to a coherent evidence subspace and avoiding noisy information.
To support multi-hop reasoning, MoG further performs an additional expert-to-expert activation round driven by entities retrieved from the activated experts. 
Experiments on three multi-hop QA benchmarks confirm MoG's effectiveness, with over 20\% relative improvements on the most challenging benchmark, MuSiQue. 
Our contributions are summarized as follows:
\begin{itemize}[leftmargin=*]
\item We propose MoG, a new GraphRAG paradigm that organizes knowledge into hub graphs with general knowledge and centrality information, along with sparsely activated domain-specific expert graphs, effectively shrinking the retrieval space to enable robust complex reasoning.
\item We introduce a topology-aware routing mechanism designed to retrieve query-specific evidence by conditionally activating the relevant expert graphs. In addition to hub-to-expert activation, the router further performs an expert-to-expert activation round to support multi-hop reasoning.
\item Through extensive experiments, we demonstrate that conditional retrieval significantly improves performance on challenging multi-hop reasoning tasks. Notably, on the challenging MuSiQue, MoG achieves over 20\% relative improvements compared to state-of-the-art methods.
\end{itemize}

\begin{figure}[!t]
	\centering
        \includegraphics[width=1.0\linewidth]{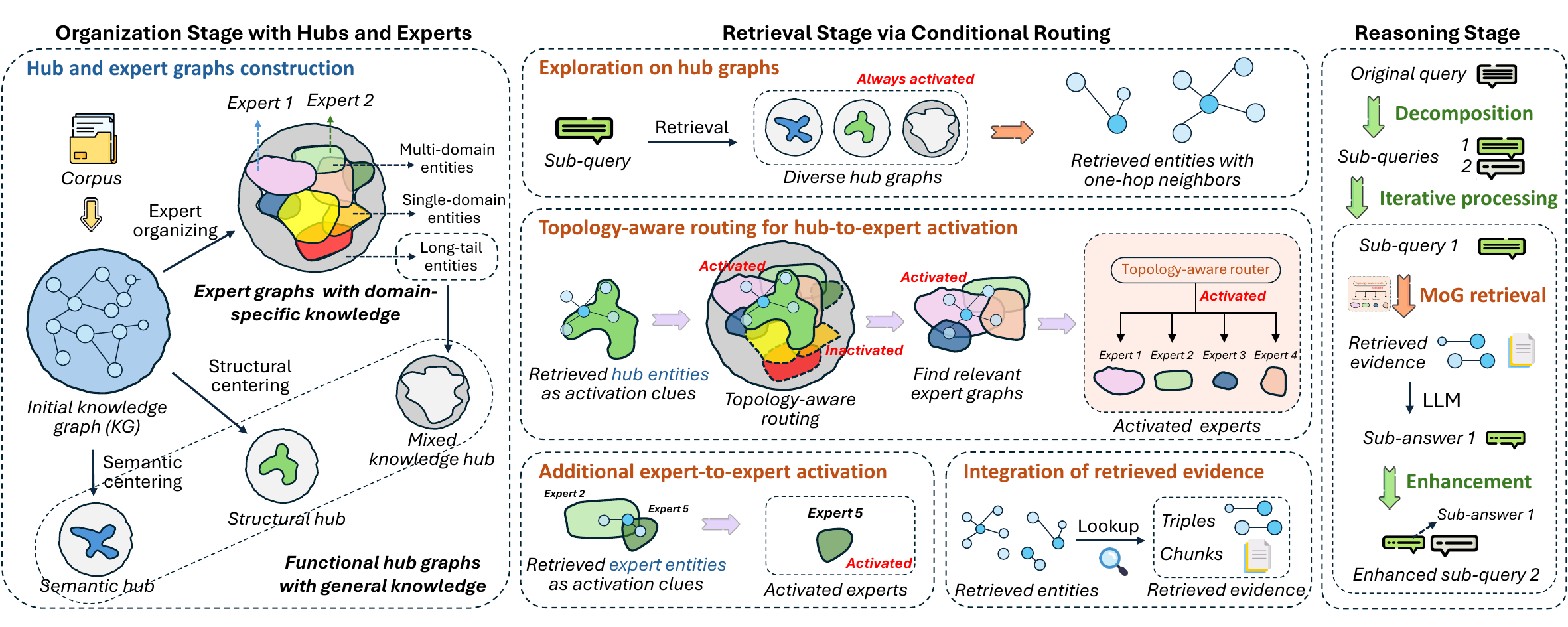}
        \caption{
        Overall framework of the proposed Mixture of Experts for Graph-based Retrieval-Augmented Generation (MoG). \textbf{(i) Organization stage with hubs and experts:} MoG organizes a constructed knowledge graph offline into overlapping fuzzy-clustered expert graphs together with complementary hub graphs (semantic, structural, and mixed) that provide globally informative entry points. \textbf{(ii) Retrieval stage via conditional routing:} Given a query, MoG first explores the hub graphs to collect clue entities and then uses a topology-aware router to sparsely activate a small subset of expert graphs for focused retrieval; additionally, MoG performs a second expert-to-expert activation to support longer reasoning chains. \textbf{(iii) Reasoning stage with iterative sub-question enhancement:} MoG iteratively enhances each decomposed sub-query with previously obtained intermediate answers, and synthesizes the accumulated retrieved evidence to generate the final answer.
        } 
        \label{fig: main figure}
\end{figure}

\section{The Framework of MoG}
In this section, we present the core idea of MoG with a coherent three-stage pipeline for effective conditional reasoning. \((i)\) In the first organization stage (Section~\ref{seE:Knowledge Organization}), MoG explicitly organizes knowledge into expert and hub graphs; \((ii)\) Then we employ a topology-aware, non-learnable router in the retrieval stage to sparsely activate only a small subset of expert graphs for each query in Section~\ref{seE: Knowledge Retrieval}; \((iii)\) Finally, the reasoning stage synthesizes the retrieved evidence for multi-hop question answering in Section~\ref{seE: Answer Generation}. We illustrate the overall framework of MoG in Figure~\ref{fig: main figure}.

\subsection{Organization Stage with Hubs and Experts}
\label{seE:Knowledge Organization}
MoG organizes knowledge in two stages. It first constructs an initial knowledge graph (KG) from the external corpus, then decomposes it into two complementary structural views for retrieval: overlapping domain-specific expert graphs, and hub graphs that capture broadly knowledge. During inference, only expert activation is query-conditioned; the graph structures and embeddings are pre-computed offline, enabling adaptive routing without the cost of online re-organization.

\paragraph{Knowledge graph construction.} Given the corpus \(\mathcal{C}\), we employ an LLM-based extraction pipeline following~\cite{dong2025youtu} to identify entities and relations and construct a KG \(G = (V, E)\), where \(V\) is the set of entities and \(E\) is the set of edges (triples) describing relations between them. Each entity \(v \in V\) is associated with one or more originating text chunks in \(\mathcal{C}\). After constructing the KG, we compute semantic representations for both entities and triples. Specifically, we use an embedding function \(f_{\mathrm{emb}} : V \rightarrow \mathbb{R}^d\) that maps each entity \(v \in V\) and triple \(e \in E\) to a \(d-\)dimensional embedding \(f_{\mathrm{emb}}(v)\) and \(f_{\mathrm{emb}}(e)\). These embeddings are computed once offline based on the constructed KG and then inherited by the corresponding expert and hub graphs according to graph membership.

\paragraph{Expert graphs via fuzzy clustering.}\label{seE:Expert graphs via fuzzy clustering.}
To capture domain-specific substructures, MoG derives a set of \(M\) expert graphs \(\{G^{\mathrm{exp}}_m\}_{m=1}^M\) by applying fuzzy \(c\)-means clustering~\cite{ghosh2013comparative} to the entity embeddings \(\{f_{\mathrm{emb}}(v)\}_{v \in V}\). The clustering produces a soft assignment \(\mu_{v,m} \in [0,1]\) for each entity \(v \in V\) and expert index \(m\), measuring the degree to which \(v\) belongs to expert \(m\). Formally, we start from an over-complete pool of \(M'\) candidate experts and solve the standard fuzzy \(c\)-means objective:
\begin{equation}
     \min_{\boldsymbol{\mu}, \{\mathbf{c}_m\}_{m=1}^{M'}} 
      \sum_{v \in V} \sum_{m=1}^{M'} \mu_{v,m}^{\,\gamma}
      \left\| f_{\mathrm{emb}}(v) - \mathbf{c}_m \right\|_2^2, \quad \text{s.t. } \sum_{m=1}^{M'} \mu_{v,m} = 1,\quad
      \mu_{v,m} \in [0,1],\ \forall v \in V,
\label{eq:fcm_combined_aligned_two_lines_appendix}
\end{equation}
where \(\mathbf{c}_m\) is the centroid of candidate expert \(m\), \(\gamma > 1\) is the fuzziness parameter controlling the softness of membership scores, and \(M'\) is the initial number of candidate experts. In practice, we choose \(M'\) to be larger than the final number of experts and then prune unused or redundant experts after clustering, resulting in \(M < M'\) effective expert graphs.

For an entity $v$, we retain only expert memberships above a threshold \(\mathcal{M}(v) = \left\{ m \mid \mu_{v,m} \ge \tau \right\}\) to obtain interpretable expert graphs, where \(\tau \in (0,1)\) is a membership cutoff. Each expert graph \(G^{\mathrm{exp}}_m\) is then constructed from the induced subgraph over entities with \(m \in \mathcal{M}(v)\). Since the assignments are fuzzy, many entities may belong to multiple experts, leading to overlapping expert graphs that better reflect the multifaceted roles of real-world entities. The entire expert-construction procedure is performed once; during inference, MoG conditionally activates only a small subset of experts, whose union of entities forms a query-specific evidence subspace. We systematically study the impact of \((M', \gamma, \tau)\), which is detailed in Figure~\ref{fig: initial expert number} in Section~\ref{sec:ablation} and Table~\ref{tb: ablation architectural configuration} in Appendix~\ref{appendix:ablation}. We also provide both quantitative and qualitative evidence to study the semantic coherence of the constructed expert graphs, including Table~\ref{tab:within_across_cosine} in Section~\ref{sec:ablation}, as well as Table~\ref{tab:silhouette_score_accuracy} and Table~\ref{tab:expert_topics_dominant} in Appendix~\ref{appendix:ablation}.

\paragraph{Hub graphs for global coverage.}
In parallel, MoG constructs hub graphs that provide general, semantically and structurally central knowledge that frequently bridges multiple experts via multi-hop connections. These hubs are always accessible during retrieval. We instantiate three types of hub graphs: a semantic hub graph, a structural hub graph, and a mixed knowledge hub graph.

\begin{itemize}[leftmargin=*,label=\textbullet]
    \item \noindent\emph{\textbf{Semantic hub graph.}}
    To identify semantically central entities, we first compute the global mean embedding of all entities. Each entity is then scored by its cosine similarity to this global representation. The top \(P\%\) of entities with the highest similarity scores are selected as semantic hubs, representing entities that are globally representative in the embedding space. We study different semantic hub conduction strategies and report the results in Table~\ref{tab:hub_selection_strategies} in Appendix~\ref{appendix:ablation}.

    \item \noindent\emph{\textbf{Structural hub graph.}}
    To capture entities that are topologically central in the knowledge graph, we measure both degree centrality (reflecting the number of an entity's direct connections) and betweenness centrality (indicating how frequently an entity acts as a bridge along the shortest paths between others) for each entity. These two factors are combined into a single structural importance score: $ c(v) = \alpha \cdot \mathrm{deg}(v) + \beta \cdot \mathrm{bet}(v),$ where \(\alpha, \beta \ge 0\) and \(\alpha + \beta = 1\). Entities ranked in the top \(P\%\) according to \(c(v)\) are selected as structural hubs. We study the robustness of MoG with different $P$, $\alpha$ and $\beta$ in Table~\ref{tb: ablation architectural configuration} in Appendix~\ref{appendix:ablation}.
    
    \item \noindent\emph{\textbf{Mixed knowledge hub graph.}}
    To maintain coverage for entities that are not strongly associated with any expert, we include entities whose membership scores fall below a predefined threshold for all experts, ensuring that long-tail or infrequent entities remain accessible during retrieval.
\end{itemize}
Entities can simultaneously appear in multiple hubs and may belong to one or more expert graphs, reflecting the overlapping structure of real-world knowledge. We provide the performance of MoG under different hub combinations in Figure~\ref{fig:ablation_musique_hub_combinations} in Section~\ref{sec:ablation}. Additionally, We study clustering initialization for expert construction and embedding backbone in Table~\ref{tab:embedding_and_init} in Appendix~\ref{appendix:ablation}.

\subsection{Retrieval Stage via Conditional Routing}
\label{seE: Knowledge Retrieval}
Given a query \(q\), MoG performs conditional retrieval to construct a focused evidence subspace tailored to \(q\). This process combines broad exploration over hub graphs and sparse activation of expert graphs via a topology-aware router \(\mathcal{R}_{\mathrm{MoG}}\). By activating only the most relevant subset of experts and integrating them with hub-derived entities, MoG concentrates retrieval on query-relevant knowledge.

\paragraph{Exploration on hub graphs.}
Hub graphs capture semantically and structurally central knowledge, with their entities bridging multiple domains via multi-hop connections.
MoG first explores all hub graphs to gather initial evidence for \(q\). It independently retrieves the top-\(K\) entities and triples in each hub according to their embedding similarity to \(q\). We denote the union of retrieved entities and triples across all hubs by \(\mathcal{V}^{\mathrm{hub}} \subseteq V\) and \(\mathcal{E}^{\mathrm{hub}} \subseteq E\), respectively. Subsequent expert activation is guided by entities retrieved from the hub graphs and their corresponding one-hop neighboring entities:
\begin{equation}
    \mathcal{V}^{\mathrm{hub}}_{\mathrm{act}}
    = \mathcal{V}^{\mathrm{hub}} \cup \bigcup_{v \in \mathcal{V}^{\mathrm{hub}}} \mathcal{N}(v),
\end{equation}
where \(\mathcal{N}(v)\) denotes the set of one-hop neighbors of \(v\) in \(G\). This augmentation incorporates structurally adjacent entities that helps expose signals for expert activation. The resulting set \(\mathcal{V}^{\mathrm{hub}}_{\mathrm{act}}\) aggregates both directly retrieved hub entities and their structurally relevant context, and is used as the joint input to the router \(\mathcal{R}_{\mathrm{MoG}}\). We ablate the inclusion of neighbors in Figure~\ref{fig:ablation_musique_neighbor_recall} in Section~\ref{sec:ablation}.

\paragraph{Topology-aware routing for hub-to-expert activation.}
The router \(\mathcal{R}_{\mathrm{MoG}}\) uses the retrieved content from hub exploration as the activation clues to select a small subset of expert graphs for focused retrieval. Recall that \(\mathcal{V}^{\mathrm{hub}}_{\mathrm{act}}\) denotes the set of clue entities obtained from the hub graphs, and that each expert graph \(G^{\mathrm{exp}}_m = (V^{\mathrm{exp}}_m, E^{\mathrm{exp}}_m)\) is defined over an entity set \(V^{\mathrm{exp}}_m \subseteq V\). For each expert \(m \in \{1,\dots,M\}\), the router first counts the number of clue entities in \(V^{\mathrm{exp}}_m\):
\begin{equation}
    c_m = \bigl|\mathcal{V}^{\mathrm{hub}}_{\mathrm{act}} \cap V^{\mathrm{exp}}_m\bigr|,
    \quad m \in \{1,\dots,M\}.
\end{equation}
Let \(C = \sum_{j=1}^M c_j\) denote the total number of clue entities covered by all expert graphs. If \(C = 0\), no clue entity is covered by any expert, which indicates that the content retrieved from the hub graphs is sufficient for the given query and there is no need to activate expert graphs. If \(C > 0\), a focused retrieval over expert graphs is required. In this case, we define the activation score \(s_m =  \frac{c_m}{C}\) for each expert.
Intuitively, experts containing more clue entities from \(\mathcal{V}^{\mathrm{hub}}_{\mathrm{act}}\) receive higher activation scores.

The router \(\mathcal{R}_{\mathrm{MoG}}\) then activates the top-\(K_{\mathrm{exp}}\) experts among those with \(s_m > 0\), yielding an active set \(\mathcal{S}^{(1)} \subseteq \{1,\dots,M\}\).
MoG then performs retrieval in the activated expert graphs using entity and triple embeddings, following the same procedure as in the hub graphs. We set \(K_{\mathrm{exp}}=5\) in our main experiment and provide the performance of MoG under different \(K_{\mathrm{exp}}\) in Figure~\ref{fig: limit activated experts} in Section~\ref{sec:ablation}. Additionally, we study alternative routing strategies in Table~\ref{tab:routing_strategies} in Appendix~\ref{appendix:ablation}.

\paragraph{Additional expert-to-expert activation.}
To support multi-hop reasoning that requires longer inference chains, MoG performs a second round of expert activation based on entities retrieved from the initially activated experts. Concretely, the first round may surface intermediate entities that serve as bridges to additional, previously unseen domains needed for later hops. If these entities belong to previously inactive expert graphs, MoG activates these expert graphs and runs an additional round of parallel retrieval. If no new experts are activated, retrieval stops. We empirically validate this two-round design and provide a detailed comparison in Figure~\ref{fig:ablation_musique_multi_round_activation} in Section~\ref{sec:ablation}.

\paragraph{Integration of retrieved evidence.}
For a query \(q\), MoG aggregates all entities and triples retrieved from both hub and expert graphs across all activation steps. Each retrieved entity is then mapped back to its corresponding text chunks in the corpus \(\mathcal{C}\), and these chunks are combined with the retrieved triples to form the final evidence sets for \(q\), encompassing evidence from all retrieval stages.:
\[
\mathrm{Evidence}(q) = \mathrm{Chunk}(\text{entities}) \;\cup\; \mathrm{Triples}(\text{entities}),
\]
where \(\text{entities}\) and \(\text{triples}\) denote the sets of entities and triples retrieved from the graph, \(\mathrm{Chunk}(\cdot)\) maps entities to their associated text chunks in \(\mathcal{C}\), and \(\mathrm{Triples}(\cdot)\) collects the retrieved triples. We select the top-\(K\) chunks and triples based on their embedding similarity with the query. Importantly, the router \(\mathcal{R}_{\mathrm{MoG}}\) is entirely non-parameter: it relies only on the graph topology and the predefined expert graphs \(\{G^{\mathrm{exp}}_m\}_{m=1}^M\), without any learned parameters.
In this way, MoG realizes a mixture-of-experts paradigm across the space of graphs, where expert graphs are conditionally activated based on topological cues derived from the respective hub and expert explorations conditioned on the queries.

\subsection{Reasoning Stage with Sub-Query Enhancement}
\label{seE: Answer Generation}

Given a query \(q\), MoG first decomposes it into a sequence of sub-queries that reflect the underlying reasoning steps. For each sub-query, MoG performs focused retrieval by conditional activation, progressively refining intermediate results and generating an answer from the aggregated evidence.

\paragraph{Query decomposition and iterative enhancement.}
Given an original query \(q\), MoG applies an LLM-based decomposition module \(\mathcal{D}\) \cite{dong2025youtu} to derive a sequence of simpler, logically ordered sub-queries: $\{q_i\}_{i=1}^N = \mathcal{D}(q)$, where \(q_i\) is the \(i\)-th sub-query and \(N\) is the number of decomposition steps. 
For each step \(i\), MoG enhances sub-query \(q_i\) by simply placing the previously obtained sub-answer \(a_{i-1}\) of \(q_{i-1}\) at the top of \(q_i\). In this way, information from the preceding sub-query is incorporated to make \(q_i\) more specific, helping the router activate more relevant experts.

For the enhanced sub-query \(q_i\), MoG applies the conditional router \(\mathcal{R}_{\mathrm{MoG}}\) to construct a focused, sub-query-specific evidence set from hub graphs and the selectively activated expert graphs. Specifically, the evidence set includes (i) \emph{textual evidence}, corpus chunks linked to the retrieved entities, and (ii) \emph{relational evidence} in the form of triples among these entities. This evidence is then assembled into an integrated context \(\mathcal{T}^{(i)}\) associated with the sub-query \(q_i\). We empirically ablate the query decomposition and sub-query enhancement module and present the results in Table~\ref{tb: Query decomposition and sub-query enhancement} in Section~\ref{sec:ablation}.

\paragraph{Evidence accumulation and answering.}
After processing all sub-queries iteratively, MoG has accumulated a logically ordered reasoning trace that combines sub-queries, their answers, and the corresponding evidence contexts: $ \mathcal{S}_{\mathrm{reason}} = \{(q_i, a_i, \mathcal{T}^{(i)})\}_{i=1}^N.$ MoG then prompts the LLM \(\mathcal{G}\) with \(q\) and $\mathcal{S}_{\mathrm{reason}}$ to generate the final answer $a$, synthesizing evidence across all reasoning steps.

\section{Experiments}
In this section, we conduct extensive experiments to evaluate the effectiveness of MoG and provide a comprehensive analysis across different domain-specific scenarios. Notably, our experiments are designed to answer the following questions: \textbf{RQ1:} Does MoG improve retrieval and generation performance compared to existing state-of-the-art baselines? \textbf{RQ2:} How does each core component of MoG contribute to the final performance, and how robust is the framework under different settings? \textbf{RQ3:} How does MoG perform in terms of efficiency, and what insights do its case studies provide?

\begin{table}[t]
	\centering
	\caption{Experimental results on multi-hop question answering across four benchmarks in terms of top-20 accuracy. We use DeepSeek-V3 as the backbone large language model. Performance is evaluated using both LLM-based semantic accuracy (LLM-Acc) and exact-match accuracy (Match-Acc).
    The performance of \underline{the best baseline} is underlined, \textbf{the best results} among all methods are in bold, and \textcolor{green!60!black}{the relative improvements} compared with the best baseline are highlighted in green.
    }
	\label{tab:main_results}
    \resizebox{1\textwidth}{!}
    {
	\begin{tabular}{lcccccccc} 
	\toprule
  \multirow{2}{*}{\textbf{Model}}
    &\multicolumn{2}{c}{\textbf{HotpotQA}}
	& \multicolumn{2}{c}{\textbf{2Wiki}} 
	&\multicolumn{2}{c}{\textbf{MuSiQue}} &\multicolumn{2}{c}{\textbf{GraphRAG-Bench}}\\
	\cmidrule(lr){2-3} \cmidrule(lr){4-5} \cmidrule(lr){6-7}\cmidrule(lr){8-9}
	&LLM-Acc &Match-Acc &LLM-Acc &Match-Acc &LLM-Acc  &Match-Acc &LLM-Acc  &Match-Acc 
        \cr \midrule 
        Zero-shot LLM &53.7  & 48.2  &41.6  & 48.8  &25.7  & 24.9 & 70.9 &63.3 \\
        G-retriever &49.9  & 45.9  &35.8  & 45.6  &23.5  & 21.2 & 70.6 &68.2 \\
        LightRAG &71.9  & 66.2  &58.0  & 60.3 &39.0 &33.4   &70.8 &67.5  \\
     E$^{2}$GraphRAG &68.7  & 65.7  &43.2  & 50.6  &28.4  &27.5 &68.7 &65.3 \\
        RAPTOR &80.9  & 67.3  &70.1  & 72.5  &48.5  & 37.2   &73.1 &70.8 \\
        HippoRAG &81.7     & 69.0    & \underline{77.9}   &80.7 & 48.3 &\underline{41.5}   &72.9 &71.0 \\
        HippoRAG2 &81.8 & 70.2  &77.3 &\underline{82.5} &50.8& 38.3   &79.4 &75.6 \\
        Youtu-GraphRAG &\underline{83.7}  & \underline{71.6}  &72.8  & 77.6  &\underline{51.4}  & 40.5  &\underline{81.5} & \underline{77.2} \\
        \textbf{MoG (Ours)} &  \textbf{86.7} {\tiny(\textcolor{green!60!black}{$\uparrow$ 3.9\%})}& \textbf{73.6} {\tiny(\textcolor{green!60!black}{$\uparrow$ 2.8\%})}& \textbf{85.7} {\tiny(\textcolor{green!60!black}{$\uparrow$ 10.0\%})} & \textbf{83.2} {\tiny(\textcolor{green!60!black}{$\uparrow$ 0.9\%})}&  \textbf{66.0} {\tiny(\textcolor{green!60!black}{$\uparrow$ 28.4\%})} & \textbf{54.1} {\tiny(\textcolor{green!60!black}{$\uparrow$ 30.4\%})}  & \textbf{84.3} {\tiny(\textcolor{green!60!black}{$\uparrow$ 3.4 \%})} & \textbf{80.5} {\tiny(\textcolor{green!60!black}{$\uparrow$ 4.3 \%})} \\
        \bottomrule
	\end{tabular}
    }
\end{table}

\begin{table}[!t]
\centering

\begin{minipage}[t]{0.44\textwidth}
  \centering
  \captionof{table}{Retrieval noise analysis. We report recall and noise rate of the retrieval component of MoG and SOTA baselines.
  }
    \label{tab:retrieval_noise}
  \resizebox{\linewidth}{!}
{%
\begin{tabular}{llccc}
\toprule
Dataset & Metric & HippoRAG2 & \makecell{Youtu-\\GraphRAG} & MoG \\
\midrule
\multirow{2}{*}{HotpotQA}
  & Recall $\uparrow$     & 89.7 & 92.3 &\textbf{95.4} \\
  & Noise Rate $\downarrow$ & 85.7 & 83.5 & \textbf{81.3} \\
\midrule
\multirow{2}{*}{2Wiki}
  & Recall $\uparrow$     & 80.2 & 91.5 &\textbf{95.4} \\
  & Noise Rate $\downarrow$ & 84.0 & 90.7 &\textbf{80.6} \\
\midrule
\multirow{2}{*}{MuSiQue}
  & Recall $\uparrow$     & 73.2 & 81.2 & \textbf{88.7} \\
  & Noise Rate $\downarrow$ & 75.2 & 73.1 & \textbf{66.1} \\
\bottomrule
\end{tabular}%
}
\end{minipage}
\hfill
\begin{minipage}[t]{0.54\textwidth}
  \centering
  \captionof{table}{LLM-Acc of methods enhanced by agentic reasoning. We apply iterative reflection-CoT (IRCoT) to the reasoning process of all methods following~\cite{dong2025youtu}, and refer to this variant as ``-IRCoT.''
  }
    \label{tb: IRCoT results}
  \resizebox{\linewidth}{!}
    {
	\begin{tabular}{lccc} 
	\toprule
        Model &  HotpotQA & 2Wiki & MuSiQue \\
        \midrule 
        RAPTOR-IRCoT  & 81.5 & 72.1 & 48.6 \\
        HippoRAG-IRCoT & 82.6 & 78.3 & 49.1 \\
        HippoRAG2-IRCoT & 83.2 & 79.9 & 51.5 \\
        Youtu-GraphRAG-IRCoT & \underline{86.5}   &\underline{85.5}  &\underline{53.6}  \\
\textbf{MoG-IRCoT (Ours)} & \textbf{87.8} {\tiny(\textcolor{green!60!black}{$\uparrow$ 1.5\%})}&\textbf{88.4} {\tiny(\textcolor{green!60!black}{$\uparrow$ 3.4\%})}  &\textbf{67.2} {\tiny(\textcolor{green!60!black}{$\uparrow$ 25.4\%})}   \\
        \bottomrule

	\end{tabular}
    }
\end{minipage}

\end{table}

\subsection{Experimental Settings}
We conduct experiments on four widely used open-source multi-hop question answering benchmarks: HotpotQA~\cite{yang2018hotpotqa}, 2Wiki~\cite{ho2020constructing}, MuSiQue~\cite{trivedi2022musique}, and GraphRAG-Bench~\cite{xiao2025graphrag}, which span diverse domains and reasoning patterns. To highlight MoG's capability for complex multi-hop reasoning, we primarily compare against advanced retrieval-augmented generation frameworks that utilize structured knowledge for multi-hop queries.
We include a zero-shot LLM baseline to quantify how much can be solved using the LLM's internal knowledge. For retrieval-based methods, we compare against a suite of representative structured or graph-based RAG methods in terms of top-20 accuracy~\cite{dong2025youtu}, including LightRAG~\cite{guo2024lightrag}, G-Retriever~\cite{he2024g}, RAPTOR~\cite{sarthi2024raptor}, E$^{2}$GraphRAG~\cite{zhao20252graphrag},  HippoRAG~\cite{jimenez2024hipporag}, HippoRAG2~\cite{gutierrez2025rag}, and Youtu-GraphRAG~\cite{dong2025youtu}. We adopt a hybrid evaluation that combines string-matching accuracy (Match-Acc) with LLM-based semantic accuracy (LLM-Acc), and we use DeepSeek-V3 as the backbone language model. A comprehensive description of implementation is provided in Appendix~\ref{appendix: experimental setting}. We also report performance with GPT-4o-mini in Appendix~\ref{Additional Main Results} for robust comparison, along with two additional baselines, GFM-RAG~\cite{luo2025gfm} and LinearRAG~\cite{zhuang2025linearrag}.

\subsection{Main Results (\textbf{RQ1})}
To address \textbf{RQ1}, we report the performance of MoG on four multi-hop QA benchmarks in Table~\ref{tab:main_results}, comparing representative graph-based retrieval-augmented generation methods tailored for multi-hop reasoning. We also evaluate MoG's ability to reduce retrieval noise, as shown in Table~\ref{tab:retrieval_noise}, and examine whether agentic reasoning can further enhance MoG in Table~\ref{tb: IRCoT results}. The key findings are as follows:

\begin{itemize}[leftmargin=*,label=\textbullet]
\item \noindent\textbf{\emph{MoG achieves state-of-the-art performance.}}
Across all datasets and under both metrics, MoG consistently attains the best results, outperforming all baselines with substantial improvement. By organizing a broad corpus into (i) hub graphs that capture general knowledge and cross-domain connections and (ii) sparsely activated expert graphs that encode domain-specific evidence, MoG partitions knowledge into smaller, coherent regions. The router then conditionally activates only the most relevant regions per query, enabling focused retrieval for multi-hop questions.

\item \noindent\textbf{\emph{MoG excels on complex reasoning.}}
On the most challenging benchmark, MuSiQue, MoG achieves more than 20\% relative improvement over the strongest baseline, highlighting the effectiveness of topology-aware routing and conditional expert activation for complex reasoning.
By conditionally activating only the most relevant expert graphs in the hop-to-expert activation round and propagating activation to previously unseen experts when multi-hop paths demand it, MoG retrieves precise evidence while capturing latent cross-domain multi-hop relationships across expert graphs.

\item \noindent\textbf{\emph{MoG's conditional activation can reduce retrieval noise.}}
Beyond end-to-end accuracy, we quantify the proportion of irrelevant evidence using recall and noise rate. Table~\ref{tab:retrieval_noise} shows that MoG achieves strong recall while reducing noise compared to baselines, suggesting that conditional activation of expert graphs improves the precision of retrieved evidence for multi-hop reasoning.

\item \noindent\textbf{\emph{Agentic reasoning further enhances MoG.}}
To study MoG with agentic reasoning method, we introduce a variant called MoG-IRCoT, which integrates the iterative reflection chain-of-thought (IRCoT) method into the MoG, following the same setting as~\cite{dong2025youtu}. We apply the same component to SOTA baselines and the results are shown in Table~\ref{tb: IRCoT results}. Equipped with the agentic reasoning module, MoG-IRCoT achieves the best performance among all models. This approach strengthens the reasoning chain and mitigates evidence omission in multi-hop reasoning tasks by iteratively applying conditional retrieval to new agent-generated sub-queries within the base MoG framework. We provide a failure case analysis and further discuss agentic reasoning in Table~\ref{tab:failure_case_analysis} in Section~\ref{seE:Model Studies}. 
\end{itemize}

\begin{figure}[htbp]
    \centering
    \begin{subfigure}{0.4\linewidth}
        \includegraphics[width=\linewidth]{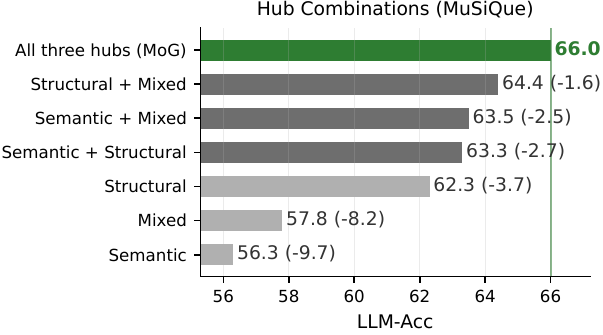}
        \caption{Ablation on hub combinations.}
        \label{fig:ablation_musique_hub_combinations}
    \end{subfigure}
    \hfill
    \begin{subfigure}{0.29\linewidth}
        \includegraphics[width=\linewidth]{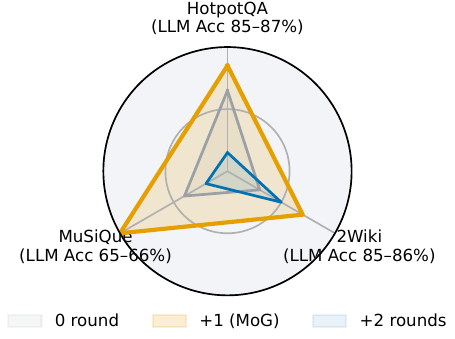}
        \caption{Ablation on multi-round expert activation during routing.}
        \label{fig:ablation_musique_multi_round_activation}
    \end{subfigure}
    \hfill
    \begin{subfigure}{0.29\linewidth}
        \includegraphics[width=\linewidth]{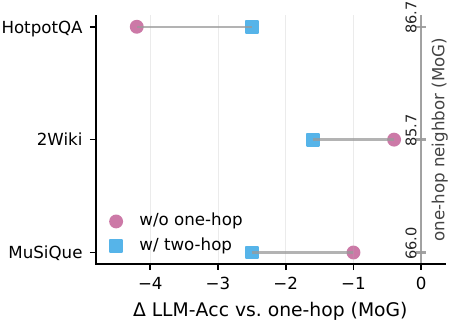}
        \caption{Ablation on neighbor recall.}
        \label{fig:ablation_musique_neighbor_recall}
    \end{subfigure}
    \caption{
    (a) Ablation on hub combinations: compare the full three-hub setup (semantic/structural/mixed) against single-hub and dual-hub variants for initial retrieval.
    (b) Ablation on multi-round expert activation during routing: vary the number of expert-to-expert activation rounds after the initial hub-to-expert retrieval.
    (c) Ablation on neighbor recall: toggle whether one-hop KG neighbors are included during retrieval, and extend the recall range to two-hop neighbors.
}
\end{figure}

\subsection{Ablation Studies (\textbf{RQ2})}
\label{sec:ablation}
To address \textbf{RQ2}, we conduct ablation studies to examine the key components of MoG and its robustness. We further analyze organization-stage hyperparameters, including the hub-size ratio \(P\%\), the weighting coefficients \(\alpha\) and \(\beta\), and the fuzzy \(c\)-means parameters \(\gamma\) and \(\tau\), in Appendix~\ref{appendix:ablation}.

\paragraph{Effect of hub combinations.}
Figure~\ref{fig:ablation_musique_hub_combinations} compares the full three-hub configuration (semantic, structural, mixed) with single- and dual-hub variants. Using all three hubs performs best, indicating that complementary semantic, structural, and long-tail views are crucial for robust initial retrieval. Among single hubs, the structural hub is the strongest, which highlights the value of topology-aware signals, while the mixed hub effectively helps recover long-tail entities. The semantic hub alone is relatively weaker but still improves robustness when combined with the others. These results validate MoG's hub design, confirming that each hub type contributes a distinct and complementary signal.

\paragraph{Single- vs.\ multi-round expert activation.}
Figure~\ref{fig:ablation_musique_multi_round_activation} evaluates how many rounds of expert-to-expert activation are performed. Adding one additional round, where newly retrieved entities from the first experts can activate further experts, consistently improves accuracy over a single hub-to-expert round by enabling longer reasoning chains. A second additional round, however, slightly degrades performance, as it mainly brings in peripheral experts and increases retrieval noise.

\paragraph{Recall of entity neighbors in retrieval.}
In MoG's retrieval stage, we not only retrieve the relevant entities but also include their one-hop neighbors in the KG.
As shown in Figure~\ref{fig:ablation_musique_neighbor_recall}, disabling the recall of one-hop neighbors leads to a noticeable performance drop,
\begin{wraptable}{r}{0.28\textwidth}
  \centering
  \begin{minipage}{0.28\textwidth}
    \centering
    \captionsetup{type=figure}
    \vspace{-4mm}
    \includegraphics[width=\linewidth]{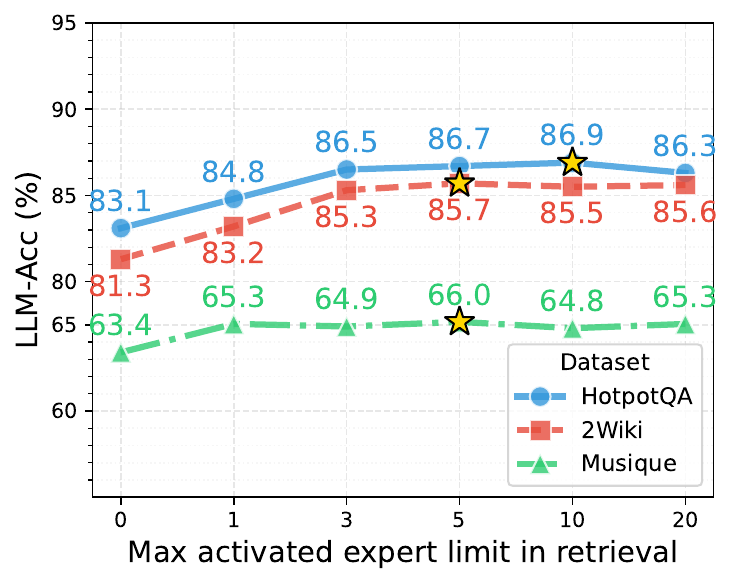}
    \caption{Effect of the upper bound on the number of activated experts in retrieval.}
    \label{fig: limit activated experts}
    \vspace{-2mm}
  \end{minipage}
  \vspace{-8mm}
\end{wraptable}indicating that multi-hop relational structure in the KG is crucial for supporting multi-step logical reasoning. We further observe that extending it to two-hop neighbors degrades performance, as it introduces weakly related context, thereby amplifying retrieval noise.

\paragraph{Varying the expert activation budget \(\boldsymbol{K_{\mathrm{exp}}}\).}
\begin{wraptable}{r}{0.5\textwidth}
  \centering
  \begin{minipage}{0.5\textwidth}
    \centering
    \captionsetup{type=table}
    \vspace{-4mm}
    \caption{Ablation studies on the impact of query decomposition with sub-query enhancement.}
\label{tb: Query decomposition and sub-query enhancement}
    \resizebox{\linewidth}{!}{
   \begin{tabular}{c |c  c  c}
   \toprule[1pt]
   Pipeline setting &  HotpotQA &2Wiki & MuSiQue \\
   \midrule
   Full pipeline  (MoG) &   \textbf{86.7}  & \textbf{85.7}  & \textbf{66.0}  	\\
   w/o iterative sub-query enhancement & 80.3 & 79.5 &  55.3	\\
   w/o query decomposition & 77.8  & 59.2 &  43.8	\\
   \bottomrule[1pt]
   \end{tabular}}
  \end{minipage}
  \vspace{-4mm}
\end{wraptable}
Figure~\ref{fig: limit activated experts} varies the maximum number of experts activated per query. Setting \(K_{\mathrm{exp}} = 0\) (using only hubs) leads to a clear accuracy drop yet remains competitive, showing that hubs provide a strong backbone for simpler queries. The results show that very small expert budgets supply insufficient cross-domain evidence, while overly large budgets introduce noise from irrelevant experts. We adopt \(K_{\mathrm{exp}} = 5\) as a robust trade-off.

\paragraph{Query decomposition and sub-query enhancement.}
Table~\ref{tb: Query decomposition and sub-query enhancement} compares full MoG with variants that remove either iterative sub-query enhancement or query decomposition. Omitting sub-query enhancement causes an accuracy drop, indicating that propagating intermediate answers is particularly important for complex questions. Disabling query decomposition and directly processing the original question yields severe degradation, confirming that concise sub-queries provide clearer intent for sparse expert activation and more precise evidence retrieval, while also supporting an explicit sub-query reasoning chain.

\subsection{Model Studies (\textbf{RQ3})}
\label{seE:Model Studies}
To answer \textbf{RQ3}, we conduct model studies to qualitatively analyze MoG's behavior, focusing on cost, failure case and semantic coherence. More model studies is reported in Appendix~\ref{appendix:Additional Model Studies}.

\begin{table}[htbp]
\centering

\begin{minipage}[t]{0.40\textwidth}
  \centering
  \captionof{table}{Offline construction token cost (million tokens) for KG construction across methods. MoG has the lowest construction cost among baselines.}
\label{tab:offline_construction_cost}
  \resizebox{\linewidth}{!}{%
  \begin{tabular}{lccc}
\toprule
Method & HotpotQA & 2Wiki & MuSiQue \\
\midrule
LightRAG & 65.1 & 39.0 & 75.5 \\
G-Retriever & 13.3 & 7.8 & 14.7 \\
HippoRAG & 13.3 & 7.7 & 14.7 \\
HippoRAG2 & 12.7 & 7.4 & 14.0 \\
Youtu-GraphRAG & 9.5 & 6.1 & 9.8 \\
MoG & \textbf{9.1} & \textbf{5.9} & \textbf{9.5} \\
\bottomrule
\end{tabular}
  }
\end{minipage}
\hfill
\begin{minipage}[t]{0.58\textwidth}
  \centering
  \captionof{table}{Average time cost of MoG compared with Youtu-GraphRAG in MuSiQue dataset. ``-'' means that the method does not have the corresponding component.}
  \label{tb: time cost}
  \resizebox{\linewidth}{!}{%
    \begin{tabular}{c | c |c  c }
      \toprule[1pt]
      \multicolumn{2}{c|}{Method} & Youtu-GraphRAG&MoG \\
      \midrule
      \multirow{4}{*}{\makecell[c]{Avg. time cost\\per generation\\component}}  & Query decomposition &6.78 s &\textbf{4.05 s} \\
      &Sub-queries answering & \textbf{7.54 s}  &  7.77 s \\
      &Initial answering & 3.46 s &   \textbf{-} \\
      &Final answering & 3.41 s & \textbf{2.41 s} \\
      \midrule
      \multicolumn{2}{c|}{\textbf{Avg. time cost of a single retrieval process}} & 5.90 s &  \textbf{0.30 s} \\
      \midrule
      \multicolumn{2}{c|}{Avg. time cost of whole pipeline} & 21.19 s &  \textbf{12.79 s} \\
      \bottomrule[1pt]
    \end{tabular}%
  }
\end{minipage}

\end{table}

\paragraph{Offline construction cost.}
We compare the offline token cost of KG construction across methods. As shown in Table~\ref{tab:offline_construction_cost}, MoG's offline construction cost is low and competitive with efficient GraphRAG baselines, while avoiding additional LLM calls in its clustering and embedding steps.

\paragraph{Retrieval and generation time cost.}
We then compare MoG's efficiency with Youtu-GraphRAG on MuSiQue in Table~\ref{tb: time cost}. At the level of a single retrieval operation, MoG is substantially faster, requiring only a small fraction of the time needed by Youtu-GraphRAG. This advantage stems from the fact that MoG conducts retrieval solely over hub graphs and a few activated expert graphs. As shown in Table~\ref{tb:activated expert graphs at inference time} in Appendix~\ref{appendix:Additional Model Studies}, each expert occupies only a minor fraction of the entire KG on average. MoG's small search space per query reduces both retrieval latency and computational overhead.
For sub-query processing, MoG is slightly slower because it performs iterative sub-query retrieval and enhancement sequentially.
Overall, the end-to-end time per query is markedly lower for MoG than for Youtu-GraphRAG, showing that conditional retrieval can simultaneously improve accuracy while maintaining practical efficiency in real-world, complex reasoning applications.

\begin{table}[t]
  \centering
  \begin{minipage}[t]{0.52\textwidth}
    \centering
    \captionsetup{type=table}
    \vspace{-3mm}
    \captionof{table}{Failure case analysis: RE: router errors (activating wrong experts); CE: chunk errors (correct experts but wrong retrieved chunks); QAE: question answering errors (correct chunks but wrong answer).}
    \label{tab:failure_case_analysis}
    \resizebox{\linewidth}{!}{%
      \begin{tabular}{llcccc}
        \toprule
        Dataset & Method & Wrong cases & RE & CE & QAE \\
        \midrule
        \multirow{2}{*}{HotpotQA} & MoG      & 133 & 15 & 14 & 104 \\
                                 & MoG-IRCoT & 122 &  8 & 12 & 102 \\
        \midrule
        \multirow{2}{*}{2Wiki}    & MoG      & 143 & 23 & 19 & 101 \\
                                 & MoG-IRCoT & 116 &  9 & 12 &  95 \\
        \midrule
        \multirow{2}{*}{MuSiQue}  & MoG      & 340 & 68 & 59 & 213 \\
                                 & MoG-IRCoT & 328 & 59 & 57 & 212 \\
        \bottomrule
      \end{tabular}%
    }
    \vspace{-3mm}
  \end{minipage}\hfill
  \begin{minipage}[t]{0.46\textwidth}
    \centering
    \captionsetup{type=table}
    \vspace{-3mm}
    \captionof{table}{Quantitative semantic coherence of expert graphs measured by within-expert vs. across-expert cosine similarity of entity embeddings. The clearly significant gaps show that clustered expert graphs partition knowledge into semantic distinct regions with a clear inter-expert differentiation.}
    \label{tab:within_across_cosine}
    \resizebox{\linewidth}{!}{%
      \begin{tabular}{lccc}
        \toprule
        Dataset & Within-expert$\uparrow$ & Across-expert$\downarrow$ & Gap \\
        \midrule
        HotpotQA & 0.73$\pm$0.07 & 0.32$\pm$0.04 & 0.41 \\
        2Wiki    & 0.68$\pm$0.05 & 0.29$\pm$0.06 & 0.39 \\
        MuSiQue  & 0.71$\pm$0.06 & 0.34$\pm$0.05 & 0.37 \\
        \bottomrule
      \end{tabular}%
    }
    \vspace{-3mm}
  \end{minipage}
\end{table}

\paragraph{Failure case analysis.}
We analyze failure cases by decomposing them into three error types: RE (router errors, i.e., activating wrong experts), CE (chunk errors, i.e., correct experts but wrong retrieved chunks), and QAE (question answering errors, i.e., correct chunks but wrong answer). Notably, QAE is beyond the scope of MoG and all RAG baselines' retrieval mechanisms, as it mainly reflects the downstream LLM’s reasoning ability given the retrieved evidence. For the two retrieval-related error types, Table~\ref{tab:failure_case_analysis} shows that agentic reasoning with iterative retrieval (e.g., IRCoT) consistently reduces RE across all benchmarks. This suggests that multi-step reasoning can revise early routing decisions and mitigate potential error propagation from incorrect expert activation, while CE remains as a complementary source of retrieval failures even when routing is improved.

\paragraph{Semantic coherence of expert graphs.}
We provide both quantitative and qualitative evidence that the clustered experts in MoG correspond to coherent topical regions. We report embedding-based separation via within-expert versus across-expert cosine similarity in Table~\ref{tab:within_across_cosine}, showing that entities assigned to the same expert are substantially more semantically aligned than entities from different experts. In the Appendix, Table~\ref{tab:silhouette_score_accuracy} complements this analysis by comparing our partition against random partitions, while Table~\ref{tab:expert_topics_dominant} provides qualitative LLM summaries of dominant topics, together supporting that MoG obtains meaningfully differentiated, domain-specialized experts.

\section{Conclusion}
We presented MoG, a new GraphRAG paradigm that addresses a key limitation of prior graph-based retrievers: uniform search over a unified knowledge base can easily introduce irrelevant neighbors and noisy reasoning paths, which is particularly harmful for multi-hop tasks. MoG tackles this by organizing a constructed KG into (i) hub graphs that provide broadly useful, structurally important entry points and (ii) sparsely activated expert graphs that capture semantically domain-specific evidence subspaces. Building on these components, we introduced a non-parametric topology-aware router that conditionally selects a small set of experts based on hub-retrieved clue entities and graph topology, thereby confining retrieval to a focused evidence region while retaining global coverage through hubs. Extensive experiments on challenging multi-hop QA benchmarks demonstrate that MoG consistently outperforms strong GraphRAG baselines, validating the central premise that conditional, topology-aware expert activation is crucial for precise evidence acquisition in complex reasoning. We believe MoG offers a general and practical framework for adaptive structured retrieval for complex reasoning task. Future work will explore richer routing signals, more efficient and robust graph construction and maintenance under evolving corpora, and extending conditional expert-style retrieval to other knowledge-intensive tasks beyond multi-hop QA.

{
\bibliography{ref}
\bibliographystyle{plain}
}

\newpage

\appendix

\section{Related Work}
\subsection{Complex Reasoning of LLM}
Recent advances in LLMs have significantly improved their ability to perform complex reasoning across diverse tasks~\cite{agrawal2023can,ferrag2025llm,guo2024large,hong2025next,jin2023large,yuan2025knapsack,zhao2024retrieval}, including multi-hop question answering~\cite{hao2024llm} and mathematical problem solving~\cite{zhang2024mathverse}. Early studies primarily attribute these capabilities to implicit reasoning patterns learned during pretraining process. Such methods often tackle the complex questions through chain-of-thought prompting or step-by-step reflection~\cite{renze2024self,wei2022chain}. Subsequent work introduces structured reasoning frameworks that decompose complex queries into intermediate steps~\cite{wu2024divide}. This decomposition supports iterative planning, verification, and refinement during inference~\cite{zhou2024isr}. Typical approaches augment LLMs with external tools~\cite{yuan2025easytool} or execution environments to support structured reasoning. EURUS ~\cite{yuan2024advancingllmreasoninggeneralists} is an open-source LLM suite optimized for complex reasoning, which is trained using a large-scale alignment dataset with structured reasoing and interaction trajectories. Despite effective in some aspects, they rely heavily on the model's internal parametric knowledge and prone to hallucination when external knowledge or background information is required. Our work addresses this by grounding reasoning not just in retrieved documents, but in a structured graph that provides fine-grained, entity-relation level multi-hop evidence, with a retrieval process that is conditionally activated by the query to reduce irrelevant noise.

\subsection{Retrieval-Augmented Generation}
Retrieval-Augmented Generation enhances large language models by integrating external knowledge retrieval into answering~\cite{arslan2024survey,yang2024crag}. Early work such as REALM combines dense retrievers with sequence-to-sequence models, enabling LLMs to access background knowledge during inference~\cite{zhu2024realm}. Subsequent studies further improve retrieval quality through better embedding models, query reformulation, and multi-stage retrieval pipelines~\cite{rusum2024vector,asai2024self}. Such methods typically employ semantic similarity retrieval, where queries and documents are embedded into a shared vector space. While effective in identifying individual passages, this paradigm treats texts as isolated units and does not explicitly consider structural dependencies among them. To better capture relationships between knowledge units, GraphRAG extends current frameworks by organizing documents or extracted entities into graph structures~\cite{zhang2025survey}. By leveraging network connectivity, GraphRAG enables multi-hop reasoning and structured information aggregation beyond flat retrieval~\cite{han2025rag}. Existing variants rely on explicit knowledge graphs, where entities and relations are extracted from text and integrated into a fine-grained KG~\cite{dong2025youtu,jimenez2024hipporag,wu2024medical,zhao20252graphrag}. Concretely, several methods construct structured graphs by extracting relational triples from passages (e.g., LightRAG~\cite{guo2024lightrag}, G-Retriever~\cite{he2024g}, HippoRAG~\cite{jimenez2024hipporag}, GFM-RAG~\cite{luo2025gfm}, and HippoRAG2~\cite{gutierrez2025rag}), and GFM-RAG further incorporates a trained GNN-based retriever to reason over relevant triples~\cite{luo2025gfm}. These methods benefit from structured semantics but require relatively higher construction costs. LinearRAG~\cite{zhuang2025linearrag} provides a cost efficiency strategy to organize knowledge. Other approaches build connections directly over text chunks using similarity-based edges or hierarchical clustering techniques rather than explicit symbolic representations. RAPTOR~\cite{sarthi2024raptor} structures the corpus into a hierarchical tree for coarse-to-fine retrieval. Overall, current methods primarily emphasize semantic relevance and logical reasoning, but they remain biased towards densely connected mainstream knowledge, overlooking the inherent asymmetry between widely shared information and specialized expertise.

\subsection{Mixture of Experts}
Mixture of Experts (MoE) is a conditional computation paradigm in which a router sparsely activates a subset of expert networks for each input, enabling models to scale parameter capacity without a proportional increase in compute and reducing interference between unrelated domains~\cite{chen2022towards,dai2024deepseekmoe}. By assigning different inputs to different experts under a top-(k) sparsity constraint, MoE focuses computation on the most relevant expert pathways and improves efficiency and task-specific performance, particularly in large-scale language models~\cite{liu2024deepseek}. We transfer this principle to RAG by modeling subgraphs as non-parametric experts, thus enabling complex reasoning.

\section{Implementation Details}\label{appendix: experimental setting}

We adopt DeepSeek-V3 and GPT-4o-mini as the backbone language models across all experiments and keep prompts and procedures fixed across datasets unless otherwise stated. For knowledge graph construction, we use the sentence-transformer encoder \texttt{all-MiniLM-L6-v2} to obtain embeddings for both entities and triples. In inducing expert graphs via fuzzy clustering, we set the membership cutoff \(\tau\) to \(0.3\), the fuzziness parameter \(\gamma\) to \(1.5\), and the initial number of candidate experts \(M'\) to \(50\). For hub selection, we set the semantic hub threshold \(P\) to \(20\%\) so that entities whose semantic scores fall within the top \(20\%\) are treated as semantic hubs, and we use the same threshold \(P = 20\%\) for structural hubs. The structural importance weights are chosen as \(\alpha = 0.8\) for degree centrality and \(\beta = 0.2\) for betweenness centrality. In expert routing, the MoG router activates the top \(K_{\mathrm{exp}} = 5\) experts among those with positive routing scores and uses this active set throughout iterative reasoning. All experiments were conducted on a CPU development machine with 50 GB of RAM.

\section{Additional Main Results}\label{Additional Main Results}
Table~\ref{tab:main_results_gpt_4o_mini} presents an additional main experiment of MoG and SOTA baselines under a different backbone LLM. Specifically, we use GPT-4o-mini as the base model and follow the top-5 evaluation protocol of LinearRAG~\cite{zhuang2025linearrag}. The results show that MoG and MoG-IRCoT consistently outperform all compared methods across all three benchmarks. The results indicate that MoG’s gains are robust across backbone LLMs, evaluation settings, and stronger graph-based competitors.

\begin{table}[htbp]
	\centering
	\caption{LLM-Acc on multi-hop question answering across three benchmarks with GPT-4o-mini. We evaluate top-5 accuracy following LinearRAG~\cite{zhuang2025linearrag}.
    }
	\label{tab:main_results_gpt_4o_mini}
    \resizebox{0.7\textwidth}{!}
    {
	\begin{tabular}{lccc} 
	\toprule
       Model (GPT-4o-mini as the base LLM) & HotpotQA &2Wiki & MuSiQue\\
        \midrule 
        G-retriever &41.9      &25.7     &15.6     \\
        LightRAG &52.7     &43.3      & 27.7   \\
        E$^{2}$GraphRAG &63.9    &49.8    &26.2    \\
        RAPTOR &55.3     &43.9    &29.7     \\
        HippoRAG &59.6     &53.2     &28.8       \\
        HippoRAG2 &64.3     &55.0      &35.3      \\
        Youtu-GraphRAG&65.3    &55.4     &34.0    \\
        GFM-RAG &65.6     &56.8     &34.6     \\
        LinearRAG &\underline{66.5}     &\underline{63.7}     &\underline{37.0}      \\
        \textbf{MoG (Ours)} & \textbf{79.7}  & \textbf{79.8}   & \textbf{53.3}    \\
        \midrule 
        Youtu-GraphRAG-IRCoT &\underline{66.3}   &\underline{58.1}      &\underline{37.5}   \\
\textbf{MoG-IRCoT (Ours)} & \textbf{82.0} & \textbf{84.2}  & \textbf{55.1}  \\
        \bottomrule[1pt]
	\end{tabular}
    }
\end{table}

\section{Additional Ablation Studies}\label{appendix:ablation}

\begin{table}[!t]
   \caption{Ablation studies on the Organization stage of MoG. We present the results of the ablation experiments for the parameters related to hub graph and expert graph region granularity. The best results are bolded. The ``(MoG)'' tag indicates the setting ultimately adopted in the MoG framework.}
\label{tb: ablation architectural configuration}
   \begin{center}
   \resizebox{0.6\linewidth}{!}{
   \begin{tabular}{c |c  c  c}
   \toprule[1pt]
   Parameter setting &  HotpotQA & 2Wiki & MuSiQue \\
   \midrule
   \multicolumn{4}{c}{Hub graph region granularity} \\
   \midrule
   \rowcolor{gray!20} \multicolumn{4}{c}{ \(P\%\) which determines the size of the semantic and structural hubs.}  \\
   \(P = 10\) & 84.4 & 85.1 &  63.7	\\
   \(P = 20\)  (MoG) &   \textbf{86.7}  & 85.7  & \textbf{66.0}  	\\
   \(P = 30\)  &85.1 &\textbf{86.1}    &  63.3    \\
   \(P = 40\)  &83.8 &85.2 &  63.3   \\
   \midrule
   \rowcolor{gray!20} \multicolumn{4}{c}{Balance of degree \(\alpha\) and betweenness \(\beta\) weights in the structural hub.}  \\
   \(\alpha=1, \beta=0\)   & 85.4     & 85.3   & 64.7  	\\
   \(\alpha=0, \beta=1\)   &  84.7    &  85.1  &  63.6 	\\
   \(\alpha=0.5, \beta=0.5\)   &   85.6   & 85.0   &  65.4 	\\
   \(\alpha=0.8, \beta=0.2\) (MoG)  &   \textbf{86.7}  & \textbf{85.7 } & \textbf{66.0}  	\\
   \midrule
   \multicolumn{4}{c}{Expert graph region granularity} \\
   \midrule
   \rowcolor{gray!20} \multicolumn{4}{c}{Fuzziness parameter $\gamma  $ and membership threshold $\tau$ in fuzzy \(c\)-means clustering.}  \\
   \(\gamma=1.5, \tau=0.3\) (MoG)  &   \textbf{86.7}  & \textbf{85.7} & \textbf{66.0}  	\\
   \(\gamma=1.5, \tau=0.5\)   &  85.6 & 85.6 & 65.8  	\\
   \(\gamma=1.5, \tau=0.7\)    &  85.7   &85.5  &  64.7 	\\
   \(\gamma=1.3, \tau=0.3\)    & 86.2 & 84.8  & 64.0  	\\
   \(\gamma=1.7, \tau=0.3\)    & 85.8  & 84.1 &  63.7 	\\
   \bottomrule[1pt]
   \end{tabular}}
   \end{center}
   \end{table}

\paragraph{Semantic and structural hub size.}
We vary the fraction \(P\%\) of entities selected from the initial knowledge graph when constructing both the semantic and structural hubs in the first block of Table~\ref{tb: ablation architectural configuration}. Across all datasets, MoG performs best when each hub contains roughly \(20\text{--}30\%\) of the entities. This range strikes a balance between coverage and specialization. If the hubs are too small, they miss important global knowledge and provide only fragmentary clues, which weakens the topology-based router leads to missed expert activations as well as incomplete evidence. If the hubs are too large, they start to absorb domain-specific information that should be delegated to expert graphs, increasing noise and diluting the benefit of specialization.

\paragraph{Balancing degree and betweenness in the structural hub.}
The structural hub is built from a composite centrality score combining degree and betweenness with weights \(\alpha\) and \(\beta\). By varying \(\alpha\) and \(\beta\) in the second block of Table~\ref{tb: ablation architectural configuration}, we observe that structural hubs defined primarily by degree, with a small contribution from betweenness, yield the most stable performance. Intuitively, degree captures local connectivity, ensuring that the hub consists of highly connected entities with rich topological context. Betweenness captures an entity's role as a bridge. Incorporating betweenness lightly helps identify globally important connectors without over-emphasizing sparse bridge entities.

\begin{table}[htbp]
   \caption{Ablation studies on the impact of iterative reflection-CoT (IRCoT) in MoG-IRCoT.}
\label{tb: ablation reasoning mechanism}
   \begin{center}
   \resizebox{0.6\linewidth}{!}{
   \begin{tabular}{c |c  c  c}
   \toprule[1pt]
   Pipeline setting &  HotpotQA &2Wiki & MuSiQue \\
   \midrule
   w/o IRCoT  (MoG) &   86.7  & 85.7  & 66.0  	\\
   Maximum 1 IRCoT step   & \textbf{88.6}   & 86.8   &  66.9	\\
   Maximum 3 IRCoT steps   & 88.1   & 87.8   &  \textbf{67.8}	\\
   Maximum 5 IRCoT steps  (MoG-IRCoT) &87.8    & \textbf{88.4 }  &  67.2	\\
   \bottomrule[1pt]
   \end{tabular}}
   \end{center}
   \end{table}

\paragraph{Impact of iterative reflection-CoT in MoG-IRCoT.}
Table~\ref{tb: ablation reasoning mechanism} evaluates MoG-IRCoT, which augments MoG with IRCoT by allowing the LLM to iteratively reflect on the current reasoning state and decide whether to generate new sub-queries. Adding IRCoT consistently outperforms the base MoG, with the optimal maximum number of reflection steps varying by dataset. This pattern suggests that a moderate number of steps helps uncover missing sub-queries and evidence, especially on complex benchmarks, whereas excessive steps introduce redundant sub-queries and noisy retrieval.

\paragraph{Expert region granularity.}
\begin{wraptable}{r}{0.3\textwidth}
  \centering
  \begin{minipage}{0.3\textwidth}
    \centering
    \captionsetup{type=figure}
    \vspace{-4mm}
    \caption{Ablation study of the initially defined expert number of \(c\)-means clustering for expert construction.}
    \label{fig: initial expert number}
    \vspace{-2mm}
    \includegraphics[width=\linewidth]{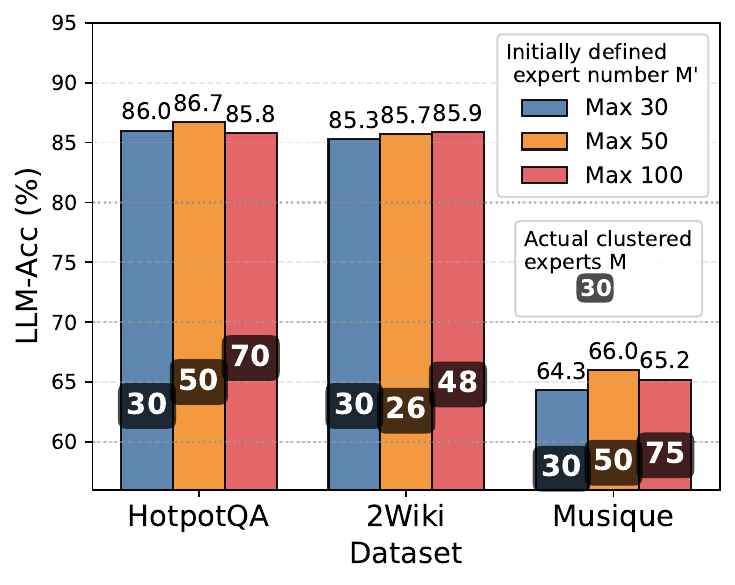}
  \end{minipage}
  \vspace{-6mm}
\end{wraptable}
For expert graphs, we study how fuzzy \(c\)-means hyperparameters affect the partitioning of expert regions in the third block of Table~\ref{tb: ablation architectural configuration}. The fuzziness parameter \(\gamma\) controls cluster assignment softness: smaller \(\gamma\) creates crisper, more distinct experts, while larger \(\gamma\) increases overlap. The membership threshold \(\tau\) determines expert assignment strictness: a higher \(\tau\) enforces domain specificity, whereas a lower \(\tau\) allows semantically related entities to be included. Given the multifaceted nature of real-world knowledge, we find a moderate \(\gamma\) with a relaxed \(\tau\) best enables controlled overlap and flexible multi-membership. We also vary the initial cluster count \(M'\), as shown in Figure~\ref{fig: initial expert number}. HotpotQA and 2Wiki are robust to \(M'\), but MuSiQue's performance declines if \(M'\) is too small or too large, likely due to reduced retrieval focus or router difficulty, respectively. Empirically, we set \(M'=50\) across all datasets as a robust configuration.

\section{Additional Model Studies}
\label{appendix:Additional Model Studies}
\paragraph{Activation of expert graphs at inference time.}
We first examine how many expert graphs are actually used during inference in Table~\ref{tb:activated expert graphs at inference time}. Although MoG constructs dozens of experts (e.g., \(50\) experts on MuSiQue), each sub-query only activates a small subset: roughly \(4\text{--}5\) experts in the hub-to-expert stage, with on average \(0\text{--}1\) additional experts brought in by expert-to-expert activation. Thus, MoG operates with a sparse, focused expert set for each sub-query rather than exploring the entire corpus. At the query level, decomposition produces about two to three sub-queries, and the total number of experts activated per query is much larger than for any single sub-query, particularly on MuSiQue. This gap shows that different sub-queries tend to route to different expert regions, reflecting the need for distinct domain knowledge in complex, cross-domain questions. These observations collectively validate MoG's conditional retrieval: distinct sub-queries require different domain information, and the topology-aware router can accurately locate the relevant experts without over-activating irrelevant ones.

\begin{table}[htbp]
\centering
\caption{The number of activated expert graphs at inference time.}
  \label{tb:activated expert graphs at inference time}
   \resizebox{0.7\linewidth}{!}{
    \begin{tabular}{c |c  c  c}
      \toprule[1pt]
      Pipeline setting &  HotpotQA &2Wiki & MuSiQue \\
      \midrule
      Number of expert graphs & 50   & 26 & 50 \\
      Avg. expert graph size (proportion of KG) &   4.6 \%  & 8.0 \%  & 3.3 \%\\
      \midrule
      \rowcolor{gray!20} \multicolumn{4}{c}{Per query.}  \\
      Avg. sub-query &2.20  &2.51  & 2.41 \\
      Avg. activated experts during sub-query processing&8.44 &7.94  & 9.22 \\
      \rowcolor{gray!20} \multicolumn{4}{c}{Per sub-query.}  \\
      Avg. activated experts in hub-to-expert activation&4.92 & 4.01 & 4.92 \\
      Avg. activated experts in expert-to-expert activation &0.83 &0.85  & 0.68 \\
      \midrule
      \bottomrule[1pt]
    \end{tabular}}%
\end{table}

\paragraph{Semantic coherence of expert graphs.}
Table~\ref{tab:silhouette_score_accuracy} offers a clustering-quality view that is consistent with the embedding-similarity evidence in Table~\ref{tab:within_across_cosine} in the main paper: MoG’s expert partition forms clearly separated semantic groups compared to non-structured assignments, and this improved separation aligns with stronger downstream QA behavior. Table~\ref{tab:expert_topics_dominant} further strengthens the interpretation by showing that different experts concentrate on distinct, human-interpretable themes across datasets, indicating that MoG discovers domain-specific expert subgraphs rather than producing interchangeable or mixed-topic partitions.

\begin{table}[htbp]
\centering
\caption{Silhouette score of expert partitioning compared to random partitions, alongside downstream QA accuracy. Higher silhouette correlates with substantially improved accuracy.}
\label{tab:silhouette_score_accuracy}
\resizebox{0.55\linewidth}{!}{%
\begin{tabular}{lcccc}
\toprule
Dataset & Partition & Silhouette Score & Accuracy \\
\midrule
\multirow{2}{*}{HotpotQA} & Ours & 0.48 & 86.7 \\
 & Random Partition & 0.05 & 32.3 \\
\multirow{2}{*}{2Wiki} & Ours & 0.39 & 85.7 \\
 & Random Partition & 0.03 & 34.7 \\
\multirow{2}{*}{MuSiQue} & Ours & 0.42 & 66.0 \\
 & Random Partition & 0.04 & 20.2 \\
\bottomrule
\end{tabular}}
\end{table}

\begin{table}[htbp]
\centering
\caption{Qualitative semantic coherence of expert graphs: LLM-summarized dominant topics for several experts across datasets, illustrating clear topical differentiation among experts.}
\label{tab:expert_topics_dominant}
\resizebox{0.7\linewidth}{!}{%
\begin{tabular}{lccc}
\toprule
Expert ID & HotpotQA & 2Wiki & MuSiQue \\
\midrule
E1 & Hollywood films & International sports & Dutch colonialism \\
E2 & British history & Frankish Empire & Music \\
E3 & Music magazines & Hip-hop & U.S.-Canada geography \\
E4 & Teen cinemas & Indian Bollywood & WWII \\
E5 & East Asian finance & Olympic athletes & Global economics \\
\bottomrule
\end{tabular}}
\end{table}

\paragraph{Additional architectural studies.}
We provide additional evidence on the robustness of MoG's architectural choices. First, we test alternative strategies for selecting semantic hubs. Table~\ref{tab:hub_selection_strategies} shows that mean-similarity hub selection performs on par with medoid-based selection and a more costly semantic PageRank alternative, suggesting that the global-mean embedding provides a near-optimal approximation of semantic centrality for identifying cross-domain connectors.
Second, we examine the sensitivity of (i) clustering initialization and (ii) embedding backbone. Table~\ref{tab:embedding_and_init} indicates that the performance is stable across different initializations, while stronger embedding models (higher-dimensional, higher-capacity encoders) bring consistent gains. This supports the claim that expert partitioning is robust to stochasticity yet can benefit from improved representation quality.

\begin{table}[htbp]
\centering
\caption{Ablation on semantic hub conduction. Mean-similarity hub selection performs comparably to medoid-based and semantic PageRank (kNN) alternatives across three multi-hop QA benchmarks.}
\label{tab:hub_selection_strategies}
\resizebox{0.6\linewidth}{!}{%
\begin{tabular}{lccc}
\toprule
Variant & HotpotQA & 2Wiki & MuSiQue \\
\midrule
Mean Similarity (Ours) & 86.7 & 85.7 & 66.0 \\
Medoid-based & 86.7 & 85.5 & 65.9 \\
Semantic PageRank (kNN) & 86.9 & 85.6 & 66.0 \\
\bottomrule
\end{tabular}}
\end{table}

\begin{table}[htbp]
\centering
\caption{Effects of embedding backbones and clustering initialization on MoG performance. Results show robustness to initialization and benefits from higher-capacity embedding models.}
\label{tab:embedding_and_init}
\resizebox{0.7\linewidth}{!}{%
\begin{tabular}{lccc}
\toprule
Variant & HotpotQA & 2Wiki & MuSiQue \\
\midrule
MiniLM-L6-v2 (384d) + Random init (Ours) & 86.7 & 85.7 & 66.0 \\
K-means++ init & 86.7 & 85.7 & 66.1 \\
Maximin init & 86.7 & 85.8 & 66.0 \\
all-mpnet-base-v2 (768d) & 86.7 & 86.2 & 67.1 \\
text-embedding-3-small (1536d) & 87.1 & 86.5 & 67.3 \\
\bottomrule
\end{tabular}}
\end{table}

\paragraph{Additional study on expert routing.}
We compare MoG's topology-aware router against representative alternatives, including semantic similarity routing and an LLM-based router. Table~\ref{tab:routing_strategies} shows that the topology-aware router consistently achieves the strongest performance across benchmarks. These results suggest that structural signals provide complementary cues beyond embedding similarity and help identify relevant expert graphs for multi-hop questions.

\begin{table}[htbp]
\centering
\caption{Comparison of routing strategies for expert activation. The topology-aware router consistently outperforms semantic-similarity and LLM-based routing.}
\label{tab:routing_strategies}
\resizebox{0.6\linewidth}{!}{%
\begin{tabular}{lccc}
\toprule
Routing & HotpotQA & 2Wiki & MuSiQue \\
\midrule
Topology-aware (ours) & 86.7 & 85.7 & 66.0 \\
Semantic similarity routing & 79.2 & 78.1 & 55.1 \\
LLM-based routing & 76.5 & 75.3 & 51.2 \\
\bottomrule
\end{tabular}}
\end{table}

\paragraph{Prompt strategy study.}
In our main experiments, we follow Youtu-GraphRAG and adopt a \emph{standard} prompt for all methods, where the LLM may use its parametric knowledge in addition to the retrieved evidence. To test whether MoG’s gains rely on this setting, we additionally remove the internal knowledge related prompt to evaluate methods in the \emph{evidence-only} prompt. Table~\ref{tab:prompt_strategy_ablation} shows that both methods drop from \emph{standard} to \emph{evidence-only}, but MoG remains consistently stronger than Youtu-GraphRAG under both prompt settings, suggesting that MoG’s improvements primarily stem from higher-quality conditional retrieval rather than prompt-induced reliance on internal knowledge.

\begin{table}[htbp]
\centering
\caption{Prompt strategy ablation comparing the standard prompt (allowing the LLM to use internal knowledge in addition to retrieved evidence) versus an evidence-only prompt (removing the internal knowledge related prompt).}
\label{tab:prompt_strategy_ablation}
\resizebox{0.6\linewidth}{!}{%
\begin{tabular}{lccc}
\toprule
Prompt setting & HotpotQA & 2Wiki & MuSiQue \\
\midrule
Youtu-GraphRAG (evidence-only) & 75.3 & 57.8 & 40.0 \\
Youtu-GraphRAG (standard)      & 83.7 & 72.8 & 51.4 \\
MoG (evidence-only)            & 82.7 & 82.3 & 60.9 \\
MoG (standard)                 & 86.7 & 85.7 & 66.0 \\
\bottomrule
\end{tabular}}
\end{table}

\section{Prompting Template}
\subsection{Triple Extraction for Knowledge Graph Construction}
To support structured retrieval over unstructured documents, we employ an information-extraction prompt that converts raw text into a schema-aligned knowledge graph. Concretely, the model is instructed to extract entities, their attributes, and inter-entity relations into a JSON representation, prioritizing the default structural schema of YoutuGraphRAG~\cite{dong2025youtu}. The prompt emphasizes flexibility beyond the predefined schema when necessary, while enforcing conciseness and non-redundancy in both attributes and triples. This procedure yields a high-coverage yet compact graph backbone that is directly consumable by our reasoning and retrieval components.

\begin{figure}[h]
\begin{tcolorbox}[
    colback=gray!5,
    colframe=gray!60,
    title=Triple Extraction for Knowledge Graph Construction,
    fonttitle=\bfseries,
    boxrule=0.5pt
]
\noindent\textbf{Instruction:} You are an expert information extractor and structured data organizer. Your task is to analyze the provided text and extract as many valuable entities, their attributes, and relations as possible in a structured JSON format. \\

\noindent\textbf{Requirements:} \\
Prioritize the following predefined schema for extraction; Schema (We use the default schema provided by YoutuGraphRAG~\cite{dong2025youtu}) Flexibility: If the context doesn't fit the predefined schema, extract the valuable knowledge as needed; Conciseness: The Attributes and Triples you extract should be complementary and no semantic redundancy. \\

\noindent\textbf{Output:}\\
Return only JSON (with a triple example).

\end{tcolorbox}
\end{figure}

\subsection{Query Decomposition}
As described in the main text, the query decomposition process breaks down complex queries into logical sub-questions. For clarity and reproducibility, we directly use the prompt and default structural schema introduced in YoutuGraphRAG~\cite{dong2025youtu}. We provide the prompts used for both the decomposition and subsequent sub-question answering in the following textbox. These prompts ensure that each sub-question is minimal and non-overlapping. The sub-question answering prompt further guarantees that responses are concise, factual, and conditioned on previously retrieved evidence and reasoning steps. Together, they formalize the step-wise reasoning procedure implemented in our framework.

\begin{figure}[h]
\begin{tcolorbox}[
    colback=gray!5,
    colframe=gray!60,
    title=Query Decomposition,
    fonttitle=\bfseries,
    boxrule=0.5pt
]
\noindent\textbf{Instruction:} You are a professional question decomposition expert specializing in multi-hop reasoning. Given the following schema and the question, decompose the complex question into the necessary number of focused sub-questions. \\

\noindent\textbf{Requirements:} \\
Prioritize the following predefined schema for extraction. \\
Each sub-question must be: As simple and direct as possible; Specific and focused on a single fact or relationship by identifying all entities, relationships, and reasoning steps needed; Answerable independently with the given schema; Explicitly reference entities and relations from the original question; Designed to retrieve relevant knowledge for the final answer; Non-redundant and non-overlapping with other sub-questions. \\
Schema (We use the default schema provided by ~\cite{dong2025youtu}), query and a one shot example. \\

\noindent\textbf{Output:}\\
Output sub-questions in a logical sequential order that builds toward the final answer. \\
For simple questions, return the original question as a single sub-question in a JSON array. \\
Return a JSON array of strings, each string being a sub-question. \\

\end{tcolorbox}
\end{figure}

\subsection{Sub-Question Answering with Retrieved Evidence}
After decomposition, each sub-question is answered using a dedicated factual QA prompt. The model is required to always return a direct, standalone answer, leveraging the retrieved context and previous reasoning steps whenever they are relevant. When the evidence is incomplete or weakly related, the model falls back on its own background knowledge while still producing a concise, context-compatible response. By disallowing meta-commentary and forcing answer-only outputs, this prompt ensures that intermediate answers remain clean, easily composable, and suitable for subsequent aggregation.

\begin{figure}[h]
\begin{tcolorbox}[
    colback=gray!5,
    colframe=gray!60,
    title=Answering with Retrieved Evidence for Sub-Query,
    fonttitle=\bfseries,
    boxrule=0.5pt
]
\noindent\textbf{Instruction:} You are a factual question answering assistant. Answer the question based on the provided context and previous reasoning steps.\\

\noindent\textbf{Requirements:} \\
You MUST provide a direct answer to the question, no matter what; If the context is helpful, use it to answer; If the context is not relevant or insufficient, use your own knowledge to provide the best possible answer; never say things like ``the context doesn't provide", ``information is not available", ``I cannot answer", etc. Output only the answer itself - no meta-commentary, no explanations about the context.  \\

\noindent\textbf{Output:}\\
Output only the answer. No explanations, reasoning, or additional text. Be precise and concise. If the context is not relevant, answer based on your knowledge.

\end{tcolorbox}
\end{figure}

\subsection{Final Answer Aggregation from Reasoning Chains}
Given the full chain of retrieved evidence and sub-question answers, we use a final aggregation prompt to produce the answer to the original query. This prompt conditions on the accumulated multi-hop context and instructs the model to return only the final, globally consistent answer, without exposing intermediate reasoning or sub-question details. As a result, the framework preserves a clear separation between latent reasoning chains and user-facing outputs, while ensuring that the final prediction fully integrates all previously collected information.

\begin{figure}[h]
\begin{tcolorbox}[
    colback=gray!5,
    colframe=gray!60,
    title=Answering with the Reasoning Chains of Sub-Queries for the Original Query,
    fonttitle=\bfseries,
    boxrule=0.5pt
]
\noindent\textbf{Instruction:} You are a factual question answering assistant.\\

\noindent\textbf{Requirements:} \\
Based on the comprehensive knowledge context accumulated through iterative retrieval, provide the final answer to the Original Question.\\

\noindent\textbf{Output:}\\
Output ONLY the final answer to the Original Question based on the sub-questions and retrieved information. Do not include any reasoning process, sub-question answers, or explanations.
Provide a comprehensive and accurate final answer.

\end{tcolorbox}
\end{figure}

\section{Limitations}
While MoG substantially improves evidence retrieval and reduces retrieval noise for multi-hop reasoning, it is fundamentally a retrieval-time framework that operates on top of a given LLM. As a result, although MoG is broadly applicable across LLM backbones, it does not directly enhance the model’s intrinsic parametric knowledge, reasoning capacity, or robustness beyond what is enabled by better conditioned evidence selection.  A promising direction for future work is to bring MoG's conditional specialization and topology-aware routing principles into LLM training. For example, supervised fine-tuning and related training paradigms could teach models to better manage evidence, plan tool usage, and perform structured multi-step reasoning with MoG-inspired expert selection, thereby improving agentic capabilities beyond retrieval-time gains.

\section*{Impact Statement}
This paper presents foundational research on retrieval and routing for multi-hop question answering with large language models, with the goal of improving reliability and efficiency of information access. The primary potential positive impact is enabling more accurate, context-grounded QA systems that can better support knowledge-intensive tasks (e.g., education, research assistance, and enterprise search) by reducing reliance on unsupported parametric recall and improving the use of provided evidence.

We do not foresee specific negative societal impacts that are uniquely introduced by this work beyond those already associated with general-purpose language models and retrieval systems. Our method does not release new private data, does not introduce new user-level profiling or surveillance capabilities, and is not intended for deployment in high-stakes decision making without appropriate safeguards.

\newpage



\end{document}